\documentclass[runningheads]{llncs}
\usepackage{graphicx}

\usepackage{tikz}
\usepackage{comment}
\usepackage{amsmath,amssymb} 
\usepackage{color}

\usepackage[accsupp]{axessibility}  
\usepackage{marvosym}

\usepackage{bm}
\usepackage{multirow}
\usepackage{booktabs}
\usepackage{makecell}
\usepackage{subfigure}

\usepackage[width=122mm,left=12mm,paperwidth=146mm,height=193mm,top=12mm,paperheight=217mm]{geometry}

\begin{document}
\pagestyle{headings}
\mainmatter
\def\ECCVSubNumber{999}  

\title{Locality Guidance for Improving Vision Transformers on Tiny Datasets} 

\titlerunning{Locality Guidance for Improving Vision Transformers on Tiny Datasets}
%
\author{Kehan Li\inst{1}\thanks{Equal contribution.} \and Runyi Yu\inst{1\star} \and Zhennan Wang\inst{2} \and
Li Yuan\inst{1,2}\thanks{Corresponding author: Li Yuan, Jie Chen} \and Guoli Song\inst{2} \and Jie Chen\inst{1,2\star\star}}
\authorrunning{K. Li et al.}
%
\institute{School of Electronic and Computer Engineering, Peking University, China \and
Peng Cheng Laboratory, Shenzhen, China\\
\email{yuanli-ece@pku.edu.cn, chenj@pcl.ac.cn}}


\maketitle

\begin{abstract}
While the Vision Transformer (VT) architecture is becoming trendy in computer vision, pure VT models perform poorly on tiny datasets. To address this issue, this paper proposes the locality guidance for improving the performance of VTs on tiny datasets. We first analyze that the local information, which is of great importance for understanding images, is hard to be learned with limited data due to the high flexibility and intrinsic globality of the self-attention mechanism in VTs. To facilitate local information, we realize the locality guidance for VTs by imitating the features of an already trained convolutional neural network (CNN), inspired by the built-in local-to-global hierarchy of CNN. Under our dual-task learning paradigm, the locality guidance provided by a lightweight CNN trained on low-resolution images is adequate to accelerate the convergence and improve the performance of VTs to a large extent. Therefore, our locality guidance approach is very simple and efficient, and can serve as a basic performance enhancement method for VTs on tiny datasets. Extensive experiments demonstrate that our method can significantly improve VTs when training from scratch on tiny datasets and is compatible with different kinds of VTs and datasets. For example, our proposed method can boost the performance of various VTs on tiny datasets (\emph{e.g.}, 13.07\% for DeiT, 8.98\% for T2T and 7.85\% for PVT), and enhance even stronger baseline PVTv2 by 1.86\% to 79.30\%, showing the potential of VTs on tiny datasets. The code is available at \textcolor{magenta}{\url{https://github.com/lkhl/tiny-transformers}}.
\end{abstract}

\section{Introduction}
\label{section:introduction}

Recently, models based on the self-attention mechanism have been widely used in visual tasks and demonstrated surprising performance, making it an alternative to convolution~\cite{dosovitskiy2020image,arnab2021vivit,carion2020end,xie2021segformer}. Of these models, ViT~\cite{dosovitskiy2020image} is the first full-transformer model for image classification, which can outperform CNNs when large training data is available. Based on ViT, a lot of works modify it and make it more adaptable to image data, which makes it possible for training Vision Transformer (VT) from scratch on medium-sized datasets (\emph{e.g.}, ImageNet-1K~\cite{deng2009imagenet} with 1.3 million samples)~\cite{touvron2021training,yuan2021tokens,liu2021swin,peng2021conformer,wu2021cvt}.

\begin{figure}[t]
    \centering
    \includegraphics[width=1.0\textwidth]{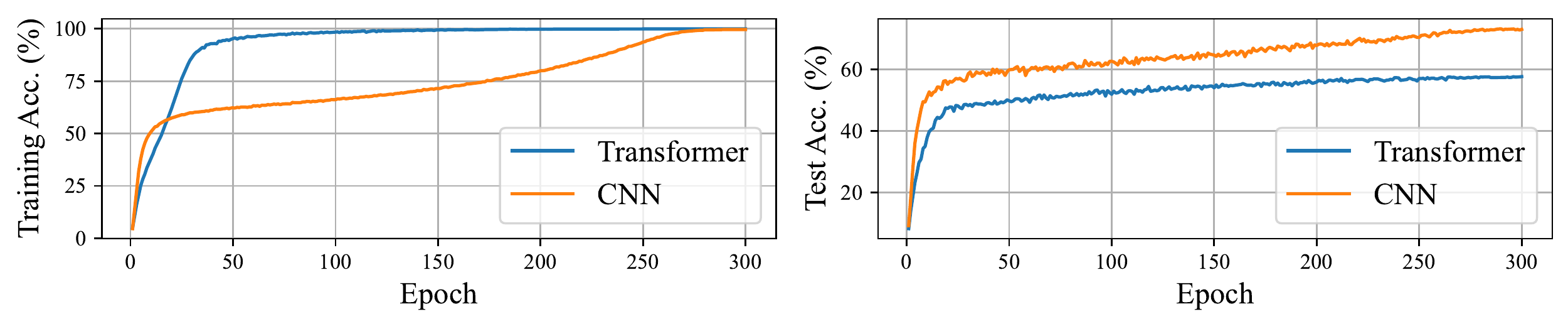}
    \caption{\textbf{Training accuracy (left) and test accuracy (right) when training CNN and Transformer on CIFAR-100 dataset.} Compared with the CNN, the Transformer fits the training set faster but has lower test accuracy, due to the difficulty of learning the local information with the globality of the self-attention mechanism.}
    \label{fig:accuracy}
\end{figure}

However, it is still difficult to train VTs from scratch on tiny datasets with a normal training policy \cite{liu2021efficient}. To be more intuitive, we train a visual transformer T2T-ViT-14~\cite{yuan2021tokens} on CIFAR-100 dataset with weak data augmentations (including padding and random cropping), where only $ 50,000 $ training samples are available. The results in Fig. \ref{fig:accuracy} show that the accuracy on the training set increases rapidly to 100\% yet the accuracy on the test set can only reach about 58\%, showing obvious overfitting. A commonly used method to address this issue is pre-training model on large datasets. However, this pretraining-finetuning paradigm has several limitations. Firstly, large-scale datasets are naturally lacking in some specific domains like medical image~\cite{zhu2021hard,zbontar2018fastMRI,menze2014multimodal}. Secondly, the model must be able to fit both the large pre-trained dataset and the small target dataset, constraining the flexibility of model designing~\cite{ke2021chextransfer}. Finally, the pre-training on a large dataset with a large model is computationally expensive. It is unacceptable that we need to retrain a new model on large dataset, even if the model architecture changes only a little, which is sometimes inevitable for specific tasks~\cite{li2018detnet,qiao2021detectors}.

\begin{figure}[t]
    \centering
    \subfigure[]{
        \begin{minipage}[b]{0.36\textwidth}
            \includegraphics[width=1\textwidth]{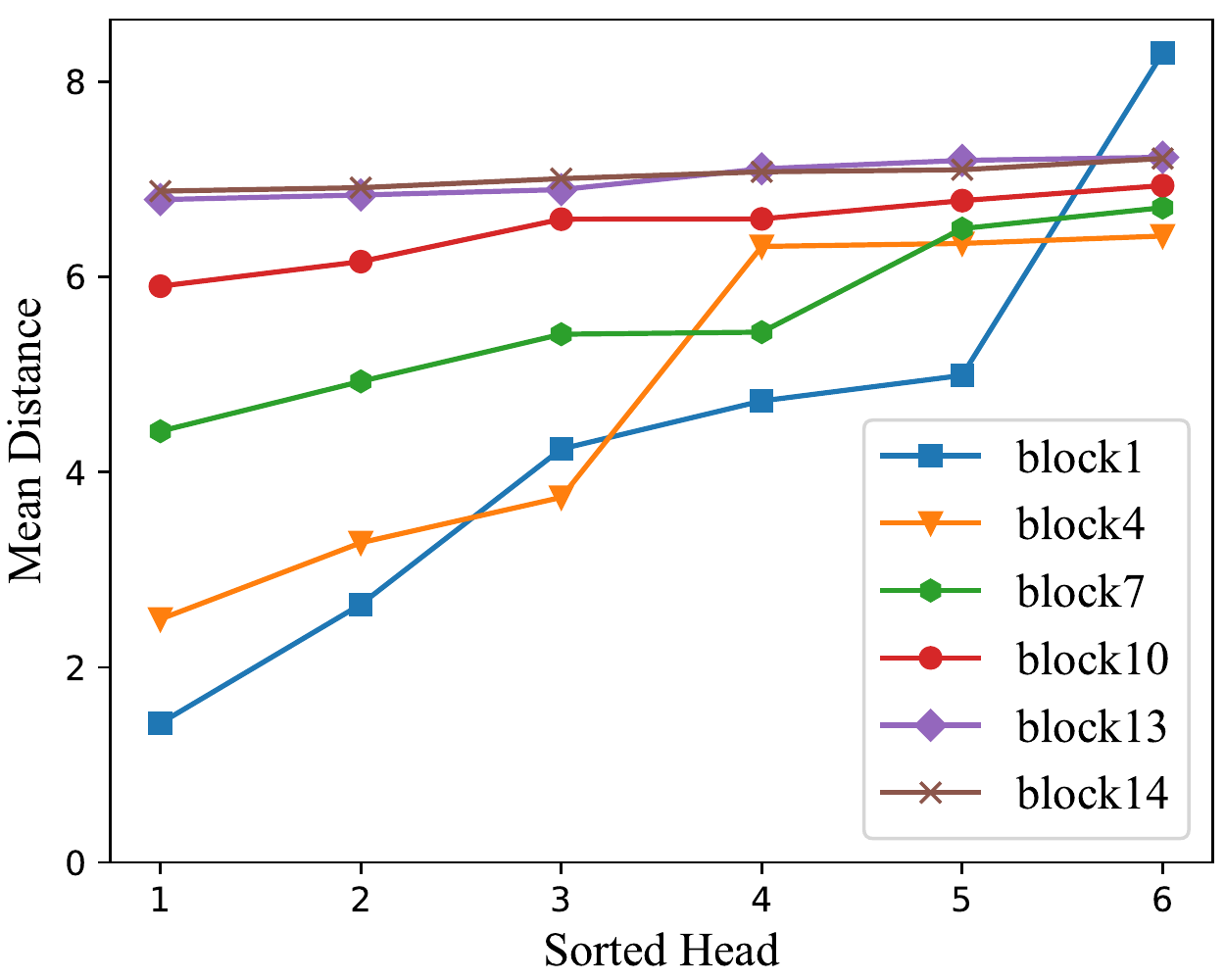} \\
            \includegraphics[width=1\textwidth]{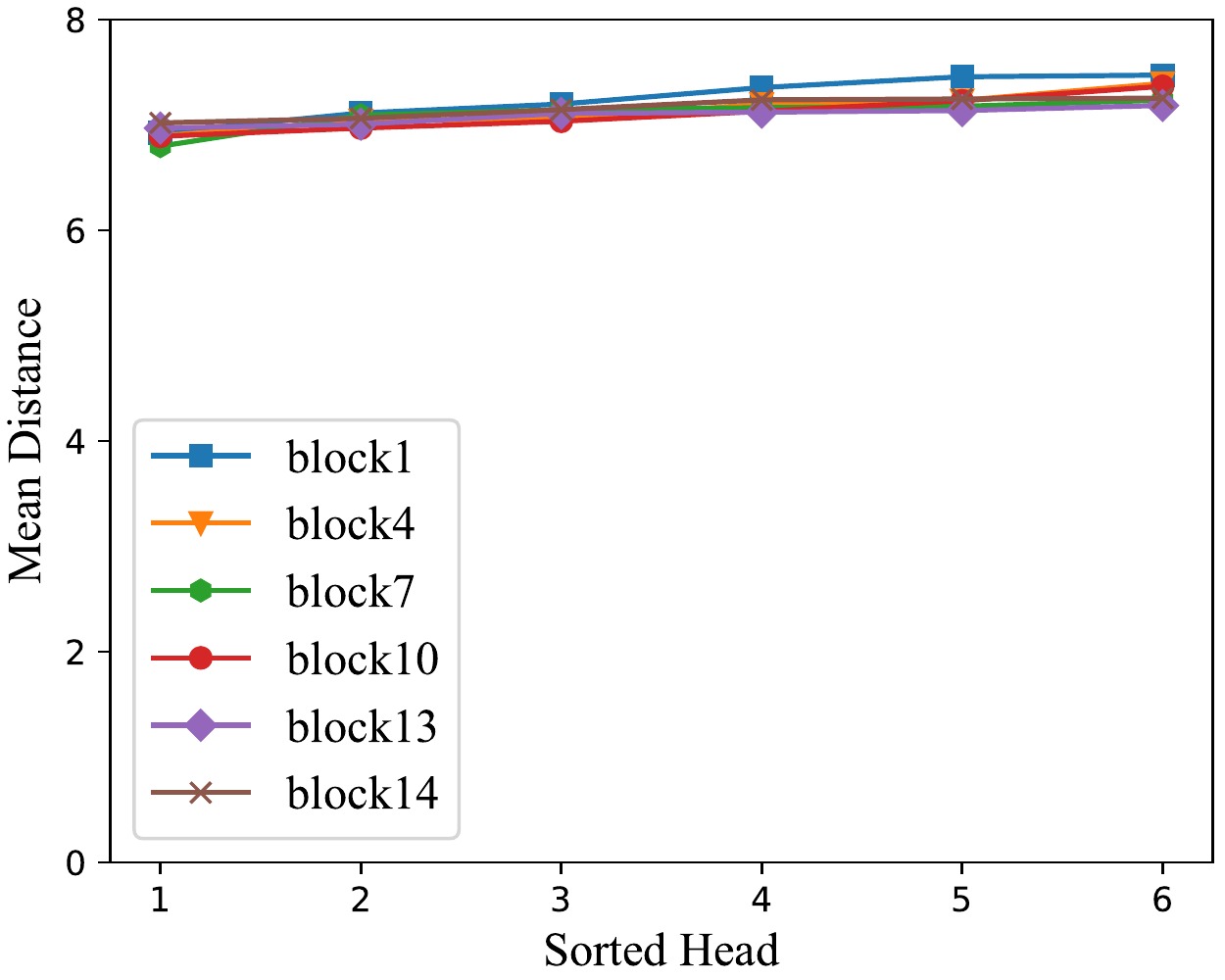}
        \end{minipage}
        \label{fig:attention_distance}
    }
    \subfigure[]{
        \begin{minipage}[b]{0.35\textwidth}
            \includegraphics[width=1\textwidth]{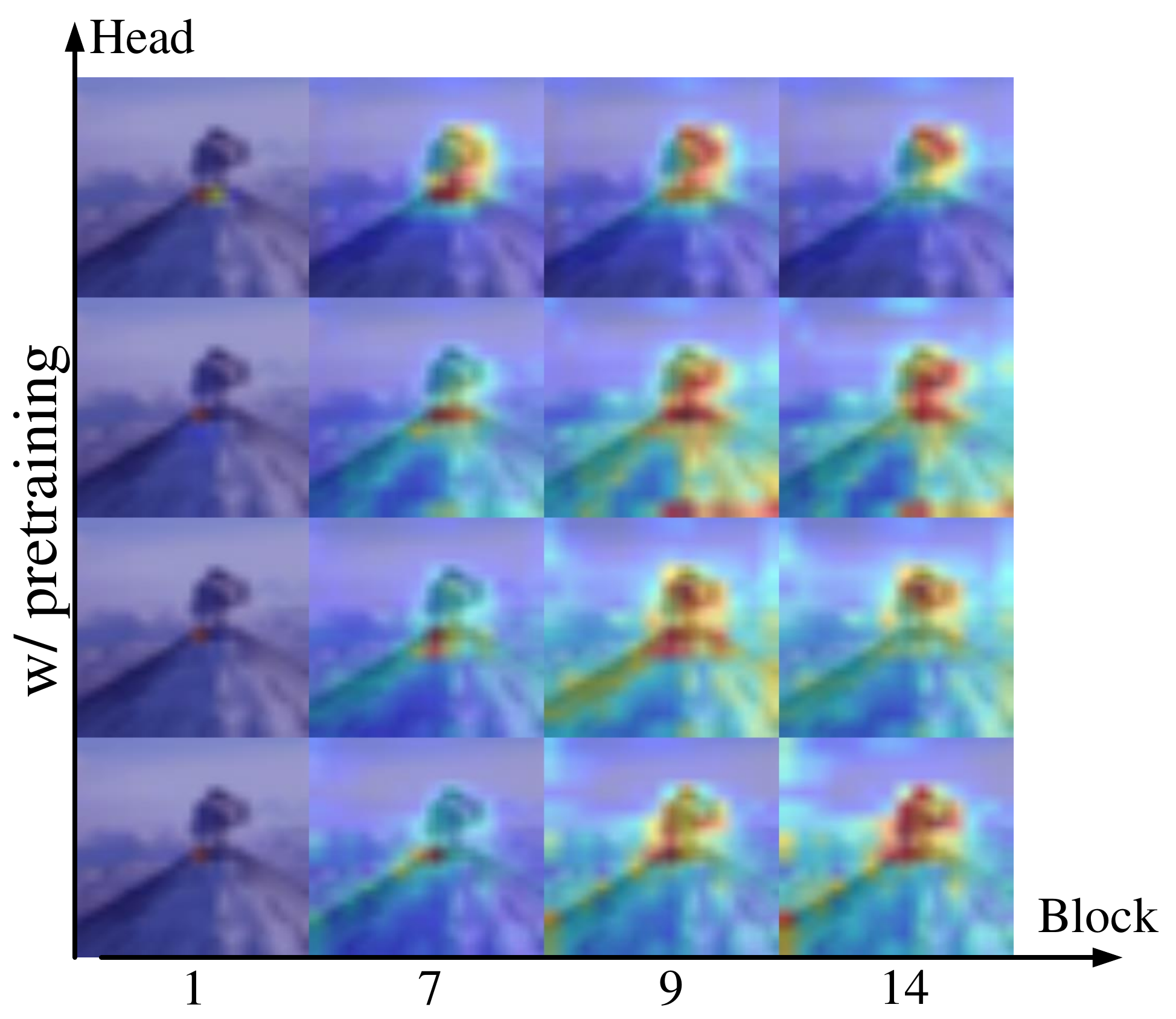} \\
            \includegraphics[width=1\textwidth]{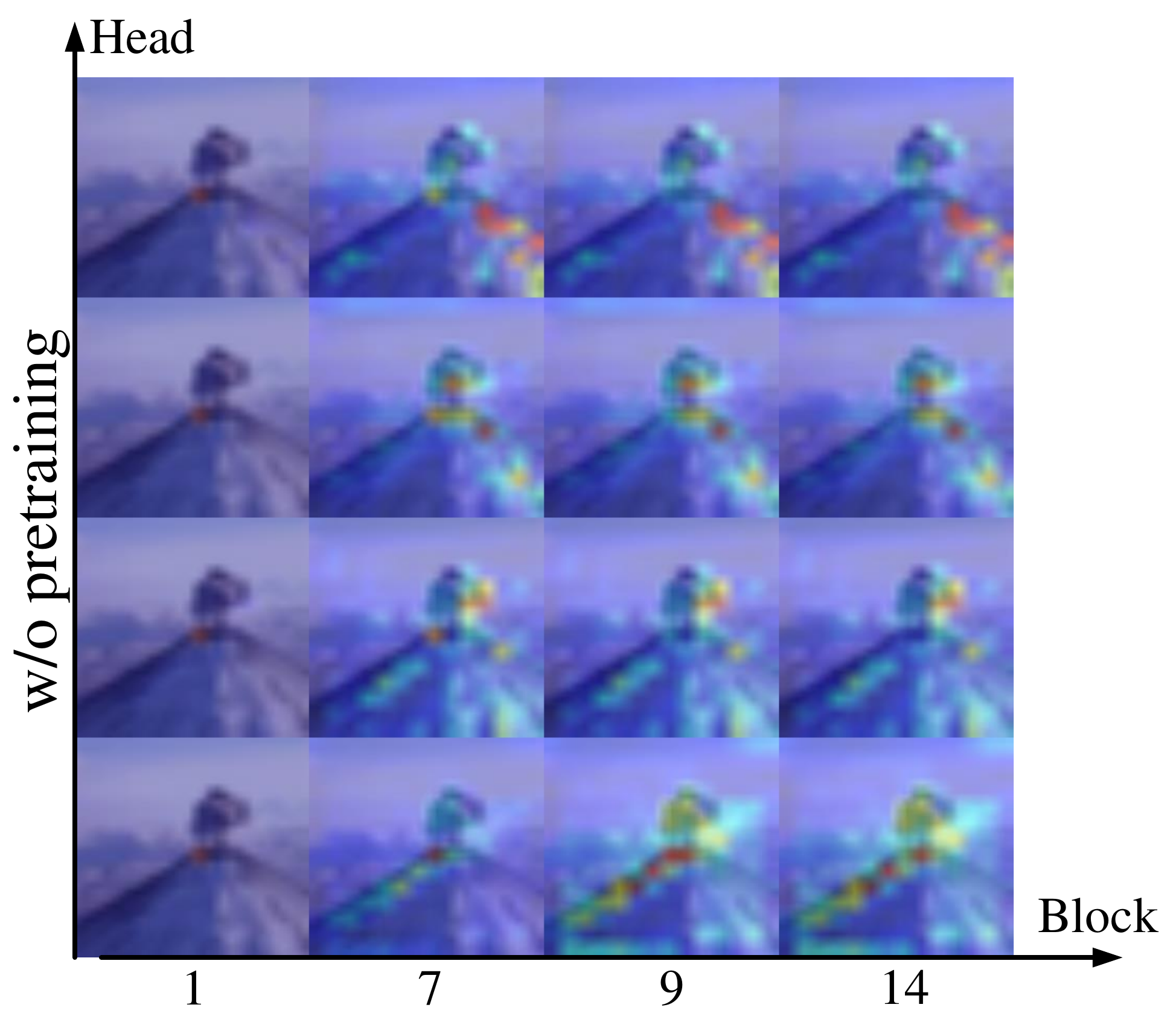}
        \end{minipage}
        \label{fig:attention_map}
    }
    \caption{\textbf{Comparison of the self-attention statistics between the model with pre-training (top) and the model without pre-training (bottom).} (a) Attention distance~\cite{dosovitskiy2020image,raghu2021vision} in different blocks. The abscissa represents the sorted attention heads. The small distance means that it is focused on the local information, and the large distance means that it is focused on the global information. (b) Self-attention map obtained by Attention Rollout~\cite{vaswani2017attention}. The columns represent sorted blocks and the rows represent sorted heads.}
    \label{fig:compare}
\end{figure}

Aiming to find a more efficient way to make VTs work well on tiny datasets, we start with analyzing why pre-training works. To do this, we compare the self-attention statistics of VTs with and without ImageNet~\cite{deng2009imagenet} pre-training. We employ the attention distance following~\cite{raghu2021vision} and the attention map by Attention Rollout~\cite{abnar2020quantifying} as the self-attention statistics, which are commonly used for analyzing self-attention mechanism~\cite{dosovitskiy2020image,raghu2021vision}. The attention distance given in Fig.~\ref{fig:attention_distance} is obtained by weighted averaging the distance between any two tokens through their attention intensity, representing the mean distance of each token to aggregate information. The attention map given in Fig.~\ref{fig:attention_map} shows the attention matrix $ q \cdot v $ of the center token. By analyzing the attention distance in Fig.~\ref{fig:attention_distance}, we find that the VT with pre-training learns to assign attention rationally. By rationally, we mean that the shallow blocks focus more on the local and the deep blocks focus more on the global. However, all blocks of the VT without pre-training only focus on the global. On the other hand, the attention map in Fig.~\ref{fig:attention_map} shows that the VT with pre-training progressively finds the relationships and finally focuses on the correct positions, while the VT without pre-training starts paying relatively fixed and uniform attention from the middle blocks. Based on these observations, we conclude that pre-training on a relatively large dataset can learn hierarchical information from locality to globality, which makes pre-trained models easier to understand images than models trained from scratch. Unfortunately, small datasets are not sufficient to extract hierarchical information for VTs.

To address these limitations, we present the locality guidance for improving the performance of VTs on tiny datasets, which helps the VTs capture the hierarchical information effectively and efficiently, as an alternative to the costly pre-training. Our proposed locality guidance is realized through the regularization provided by convolution, motivated by the inherent local-to-global hierarchy of convolutional neural network (CNN)~\cite{lee2011unsupervised}. Specifically, we employ an already trained lightweight CNN on the same dataset to distill the VT in hidden layers. Therefore, there are two tasks for the VT. One is to imitate the features generated by the CNN (\emph{i.e.}, receive the guidance), and the other is to learn by itself from the supervised information. The imitation task is auxiliary and thus does not impair the strong learning ability of VTs.

The efficiency of our method is reflected in three aspects. a) Since the feature imitation is just used as an auxiliary task to guide the VT, the performance of the CNN will not be the bottleneck for the VT, and therefore it is possible to utilize a lightweight model and low image resolution, making the computational cost of CNN as small as possible. b) Information from the CNN is only needed when training, thus there is no extra computational cost when inference. c) Our method can largely accelerate the convergence and reduce the training time of the VT.

The proposed method shows its effectiveness on various types of VT and datasets. On CIFAR-100 dataset~\cite{nilsback2008automated}, our method achieves 13.07\% improvement for the DeiT~\cite{touvron2021training} baseline and improves a stronger baseline PVTv2 by 1.86\% to 79.30\%, demonstrating the potential of using VTs on tiny datasets as the alternatives to CNN. Moreover, we adopt our method on Chaoyang dataset~\cite{zhu2021hard} and show its practicality and validity on medical imaging, where the large-scale dataset for pre-training is hard to obtain. These experiments show that our locality guidance method is generally useful and can advance the wider application of transformers in vision tasks.

\section{Related Work}

\noindent
\textbf{Vision Transformers.} Transformer, a model mainly based on self-attention mechanism, is first proposed by Vaswani et al.~\cite{vaswani2017attention} for machine translation and is widely used in natural language processing tasks~\cite{devlin2018bert,brown2020language} and cross-modal tasks~\cite{yu2019deep,yang2021tap,li2022joint}. ViT~\cite{dosovitskiy2020image} is the first pure visual transformer model to process images, and can outperform CNNs on image classification task with large-scale training data~\cite{dosovitskiy2020image}. However, when massive training data is not available, ViT can not perform well~\cite{d2021convit,liu2021efficient}. Aiming to train from scratch and surpass CNNs on medium datasets (\emph{e.g.}, ImageNet-1K~\cite{deng2009imagenet}), there are lots of improved models based on ViT, including adopting a hierarchical structure~\cite{yuan2021tokens,wang2021pyramid,liu2021swin,heo2021rethinking,zhang2021aggregating,wang2021pvtv2}, introducing inductive bias~\cite{touvron2021training,peng2021conformer,yuan2021incorporating,wu2021cvt}, performing self-attention locally~\cite{liu2021swin,zhang2021aggregating,yuan2021volo}, \emph{etc}. But for tiny datasets, most of these methods still perform poorly.
\smallskip

\noindent
\textbf{Hybrid of Convolution and Self-attention.} Introducing the convolutional inductive bias to transformers has been proved effective in visual tasks.
To make use of both the locality of convolution and the globality of self-attention, Peng et al.~\cite{peng2021conformer} build a hybrid model including a CNN branch, a transformer branch and feature coupling units. Yuan et al.~\cite{yuan2021incorporating} incorporate convolution in tokenization module and feed-forward module of transformer block, while Wu et al.~\cite{wu2021cvt} introduce convolution when embedding tokens and calculating $ q, k, v $. Unlike these methods which modify the structure of VT to incorporate convolution, we keep the pure VT structure unchanged. We just employ CNN as a regularizer to guide the feature learning of VT. Therefore, our method is very simple and easy to implement, and can be used in a plug-and-play fashion. Moreover, we also show that our method can be combined with them to further improve the performance.
\smallskip

\noindent
\textbf{Vision Transformers on Tiny Datasets.} There are only a few studies focusing on how to use VTs on tiny datasets~\cite{liu2021efficient,hassani2021escaping,touvron2021training}. Liu et al.~\cite{liu2021efficient} propose an auxiliary self-supervised task for encouraging VTs to learn spatial relations within an image, making the VT training much more robust when training data is scarce. Hassani et al.~\cite{hassani2021escaping} focus on the structure design for tiny datasets, which includes exploiting small patch size, introducing convolution in shallow layers and discarding the \emph{classification} token. We argue that exploiting small patch size will bring quadratic computational complexity increases which are unacceptable when the size of the image is large. Touvron et al.~\cite{touvron2021training} adopt a longer training schedule of 7200 epochs for the VT on CIFAR-10 dataset to obtain a good result. In contrast, our proposed locality guidance for VT achieves significant performance improvements on tiny datasets while employing only 100/300 epochs.
\smallskip

\noindent
\textbf{Knowledge Distillation.} Our method is also related to knowledge distillation, which is first proposed by Hinton et al.~\cite{hinton2015distilling} and becomes a commonly used technology for model compression and acceleration~\cite{luo2016face,he2019knowledge,wu2019distilled,wang2019distilling}. The knowledge to be distilled can be divided into three kinds~\cite{gou2021knowledge}, \emph{i.e.}, response-based knowledge~\cite{hinton2015distilling,kim2017transferring}, feature-based knowledge~\cite{romero2014fitnets,komodakis2017paying,yim2017gift,he2019knowledge} and relation-based knowledge~\cite{passalis2020heterogeneous,tung2019similarity}. Our method is highly related to feature-based knowledge distillation, or also feature imitation~\cite{wang2019distilling}, which is first defined in Fitnets~\cite{romero2014fitnets}. Following Fitnets, there are many variants of representing knowledge, \emph{e.g.}, attention map~\cite{komodakis2017paying}, truncated SVD~\cite{yim2017gift}, average pooling~\cite{changyong2019knowledge}, \emph{etc}. Most applications of knowledge distillation are based on the setting of a strong teacher model and a weak student model, to achieve model compression and acceleration. Different from them, our goal of using CNN teacher is providing the locality guidance for the VT, making the learning process on tiny datasets easier so that the VT can be trained better. In our setting, the performance of the teacher will not be the performance bottleneck of the VT, since the VT is still learning by itself and can play to the advantage of the transformer. Therefore, a lightweight CNN teacher would suffice. A recent proposed method DeiT~\cite{touvron2021training} also uses knowledge distillation, which makes the VT learn the classification results of the CNN teacher. However, a CNN of comparable size to the VT is required in DeiT. By comparison, our method can achieve much higher performance with just a lightweight CNN.

\section{Method}
\label{section:method}

In this section, we first formulate the overall training procedure of the proposed method, followed by the detailed designs of our method which consist of the guidance positions and the architecture of the guidance model.

\subsection{The Overall Approach}

\begin{figure}[t]
    \centering
    \subfigure[]{
        \begin{minipage}[b]{0.822\textwidth}
            \includegraphics[width=1\textwidth]{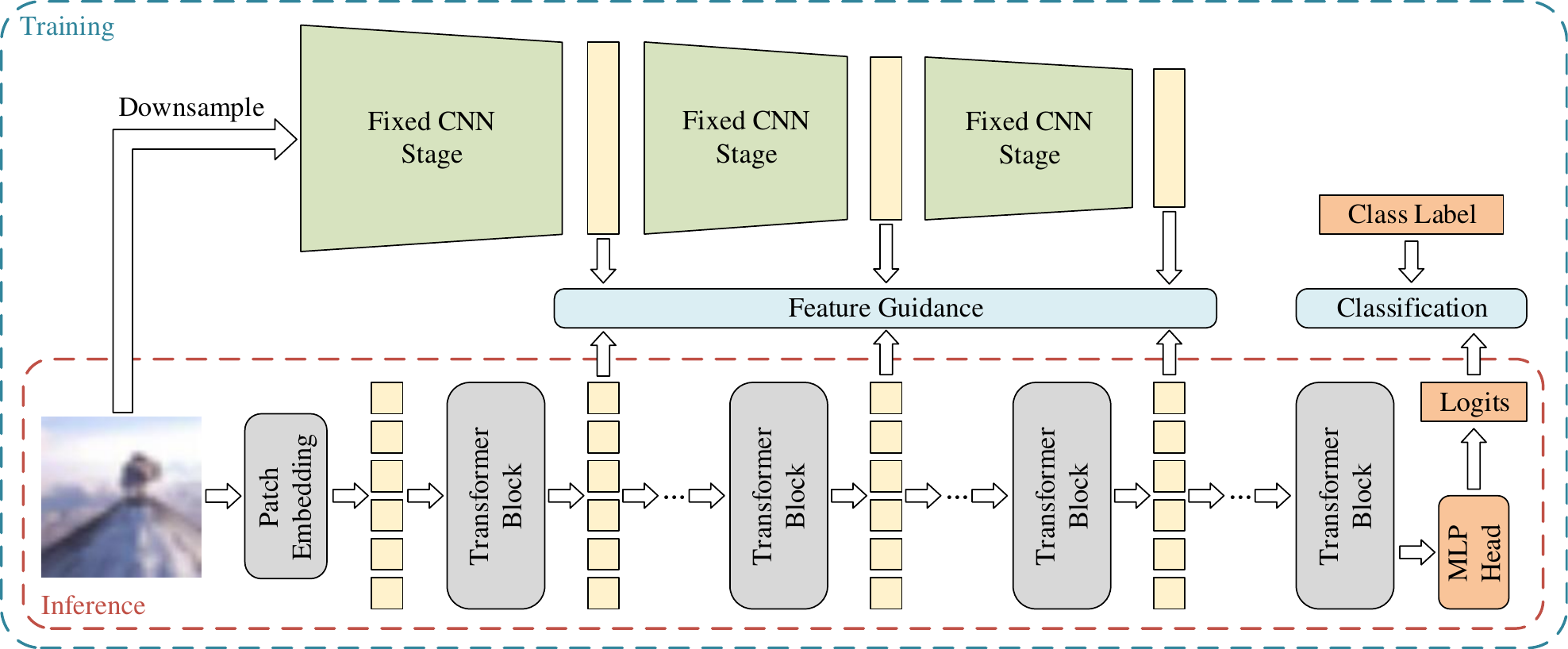}
        \end{minipage}
    }
    \subfigure[]{
        \begin{minipage}[b]{0.134\textwidth}
            \includegraphics[width=1\textwidth]{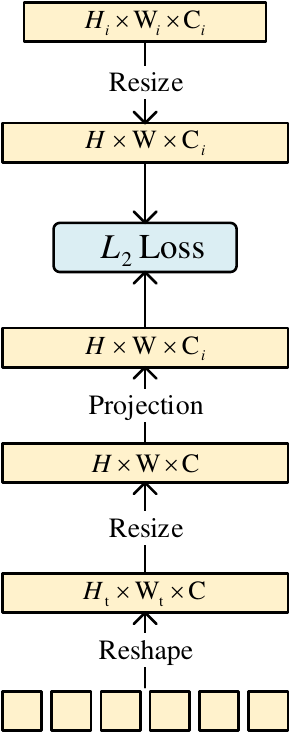}
        \end{minipage}
    }
    \caption{\textbf{Illustration of the proposed method.} (a) The process of our method. There are two tasks when training. A lightweight CNN trained on the same dataset is used to help the VT to learn local information and the VT also learns from the supervision of class labels. (b) The details of feature guidance. Transformations in both spatial and channel dimensions are performed to align features from different models and $ L_2 $ Loss is used to reduce the distance between the transformed features.}
    \label{fig:method}
\end{figure}

To improve the poor performance as well as speed up the convergence of VTs when training from scratch on tiny datasets, we propose to provide locality guidance for VTs to aid in the process of learning local information. As shown in Fig.~\ref{fig:method}, we introduce a lightweight CNN trained on the same dataset that the VT used. During training, there are mainly two tasks. Firstly, the semantic gaps between token sequences from different layers of the VT and the features from the CNN are forced to be close to some extent. This procedure is implemented by feature alignment in both spatial and channel dimensions and a feature distance metric as the loss function to be optimized, motivated by feature-based knowledge distillation~\cite{romero2014fitnets,wang2019distilling,ahn2019variational,shen2019meal,heo2019knowledge}. Secondly, the VT learns knowledge by itself through the supervision of class labels, so that it can understand the images in its own way. The proposed method is aimed at how to train a VT effectively and efficiently on tiny datasets, so we just modify the training process and there are no additional designs in the structure of the VT. In addition, the CNN is no longer needed during inference.
\smallskip

\noindent
\textbf{Feature Alignment}
A typical VT consists of two parts, the patch embedding module and a series of stacking transformer encoder blocks. The patch embedding module blocks the image and performs a linear projection to generate tokens. Each transformer encoder block contains a multi-head self-attention layer and a two-layer MLP for processing information of tokens. For the input image $ X \in \bm{R}^{H \times W \times 3} $, the information flow of VT can be formulated as
\begin{equation}
    \begin{aligned}
        & \mathcal{T}_0 = PatchEmbedding(X),\ \mathcal{T}_0 \in \bm{R}^{L \times C}, \\
        & \mathcal{T}_i = Block_i(\mathcal{T}_{i-1}),\ \mathcal{T}_i \in \bm{R}^{L \times C},
    \end{aligned}
\end{equation}
where $ \mathcal{T}_0 $ is the initial token sequence produced from the image, $ \mathcal{T}_i $ is the token sequence after transformer encoder block $ i $, $ L $ is the number of tokens and $ C $ is the embedding dimension.

A CNN is usually composed of multiple stages. As the depth increases, the resolution gradually decreases. The information flow of CNN can be formulated as
\begin{equation}
    \begin{aligned}
        & \mathcal{M}_1 = Stage_1(X),\ \mathcal{M}_1 \in \bm{R}^{H_1 \times W_1 \times C_1}, \\
        & \mathcal{M}_i = Stage_i(\mathcal{M}_{i-1}),\ \mathcal{M}_i \in \bm{R}^{H_i \times W_i \times C_i},
    \end{aligned}
\end{equation}
where $ \mathcal{M}_i $ is the feature map after stage $ i $. 

Given a token sequence $ \mathcal{T}_{i} \in \bm{R}^{L \times C} $ from the VT and a feature map $ \mathcal{M}_{j} \in \bm{R}^{H_{j} \times W_{j} \times C_{j}} $ from the CNN, due to the difference in both spatial and channel dimensions between them, they need to be transformed into the same size for the optimization convenience. We first restore the spatial dimension of features from the VT by reshaping operation, because the images are naturally two-dimensional in the spatial dimension. Then in order to calculate the distance metric more accurately, the two features are adjusted to the same size, which is the largest length and width of them. We employ the linear up-sampling to implement the resizing operation. These spatial feature alignment operations are formulated as follows
\begin{equation}
    \begin{aligned}
        & \hat{\mathcal{T}_{i}} = Reshape(\mathcal{T}_{i}),\ \hat{\mathcal{T}_{i}} \in \bm{R}^{H_t \times W_t \times C}, \\
        & \hat{H}= \max(H_t, H_{j}),\ 
        \hat{W} = \max(W_t, W_{j}), \\
        & \hat{F_{vt}} = Resize(\hat{\mathcal{T}_{i}}),\ \hat{F_{vt}} \in \bm{R}^{\hat{H} \times \hat{W} \times C}, \\
        & F_{cnn} = Resize(\mathcal{M}_{j}),\ F_{cnn} \in \bm{R}^{\hat{H} \times \hat{W} \times C_{j}},
    \end{aligned}
\end{equation}
where $ \hat{F_{vt}} $ and $ F_{cnn} $ are the spatially transformed features from the VT and the CNN, respectively.

For the alignment of channel dimension, a learnable point-wise linear projection are performed on the features from VT
\begin{equation}
    F_{vt} = Linear(\hat{F_{vt}}),\ F_{vt} \in \bm{R}^{\hat{H} \times \hat{W} \times C_{j}},
    \label{equation:linear}
\end{equation}
where $ Linear $ is the learnable linear projection, which is implemented by $ 1 \times 1 $ convolution. The linear projection acts as not only a transformation to align the channel dimension, but also a simple yet effective way to prevent the VT from learning the same features as the CNN, which may lead to a performance bottleneck. To do this, the learning of VT is flexible and the capability of the VT will not be limited by the CNN. It is also applicable to use other channel dimension transformation functions that do not align features forcibly in this framework (\emph{e.g.}, attention map~\cite{komodakis2017paying}, similarity matrix~\cite{tung2019similarity}), but the learnable linear projection is simpler and more flexible, which is shown in ablation study of Section~\ref{section:ablation}.
\smallskip

\noindent
\textbf{Dual-task Learning Paradigm}
With the two one-to-one sets of transformed features $ \{F_{vt}^{i}\,|\,i=1,2,\cdots,k\} $ and $ \{F_{cnn}^{j}\,|\,j=1,2,\cdots,k\} $, we use $ L_2 $ distance metric to realize feature guidance
\begin{equation}
    L_{guidance} = \sum_{i=1}^k \frac{1}{\hat{H_i} \cdot \hat{W_i}}||F_{vt}^{i} - F_{cnn}^{i}||_F^2.
\label{equation:loss}
\end{equation}
where $ k $ is the number of features chosen to perform guidance. Then the total loss can be formulated as
\begin{equation}
    L = L_{cls} + \beta L_{guidance},
\label{equation:total_loss}
\end{equation}
where $ L_{cls} $ is the cross-entropy loss for classification task. The final loss consists of two parts, corresponding to the two tasks, in which $ L_{cls} $ allows the VT to learn by itself while $ L_{guidance} $ forces the VT to imitate the features learned by the CNN for the purpose of incorporating local information better. Under such a dual-task setting the VT is able to express as its own way instead of just copying the features learned by the CNN, so that the performance of the CNN is not a decisive factor for the performance of VT, making it unnecessary to adopt a large-capacity CNN, which is proved by our experiments in Section~\ref{section:ablation}. The hyperparameter $ \beta $ is used to balance imitation and self-learning and we show its influence through ablation study in Section~\ref{section:ablation}.

\subsection{Guidance Positions}

We now detail the rule of constructing the two one-to-one feature sets, \emph{i.e.}, deciding the positions to perform guidance both in the VT and the CNN. The two feature sets are defined as
\begin{equation}
    \begin{aligned}
        & \bm{S_T}=\{\mathcal{T}_{i_1}, \mathcal{T}_{i_2}, \cdots, \mathcal{T}_{i_k}\},\ \mathcal{T}_{i_\cdot} \in \bm{R}^{L \times C}, \\
        & \bm{S_C}=\{\mathcal{M}_{j_1}, \mathcal{M}_{j_2}, \cdots, \mathcal{M}_{j_k}\},\ \mathcal{M}_{j_\cdot} \in \bm{R}^{H_{j_k} \times W_{j_k} \times C_{j_k}},
    \end{aligned}
\end{equation}
where $ i_1, i_2, \cdots, i_k$ and $ j_1, j_2, \cdots, j_k$ are the indexes of blocks or stages in the VT and the CNN, respectively. It is worth noting that information in one layer is produced based on previous layers in both the VT and the CNN, due to a feed-forward structure. Therefore, indexes of features in the VT and the CNN should have the same relative position (\emph{e.g.}, $ i_1, i_2, \cdots, i_k$ should be monotonically increasing if $ j_1, j_2, \cdots, j_k$ are monotonically increasing). Based on this rule, as well as making use of the information learned by the CNN as much as possible, we select the features after each stage of the CNN and the features uniformly distributed within a specific depth of the VT correspondingly to implement guidance through the loss function shown in Equation~\eqref{equation:loss}. The regulation of choosing features can be summarized as
\begin{equation}
    \begin{aligned}
        & j_k = k, \\
        & i_k = \lfloor (k-1) \cdot \frac{R \cdot N_T - 1}{N_C-1} \rfloor + 1,
    \end{aligned}
    \label{equation:position}
\end{equation}
where $ k \in \{x\,|\,1 \leq x \leq N_C, x \in \bm{Z}\} $ is the index of the selected feature. $ N_T $ and $ N_C $ are the number of blocks or stages in VT and CNN, respectively. $ R $ is a hyperparameter to control the depth of performing guidance in the VT. We provide further experimental results to compare different choices in ablation study in Section~\ref{section:ablation}.

\subsection{Architecture of The CNN}
Unlike most applications of knowledge distillation which focus on transferring the knowledge of a strong teacher model to a weak student model to realize model compression, our method aims at realizing locality guidance from the teacher, rather than totally transferring the features of the teacher. Under our framework, the VT will learn by itself through the supervision of class labels, while the CNN just provides some guidance on locality for the VT. Therefore, the weak CNN will not be a performance bottleneck for the strong VT.

In order to achieve an efficient training process, we chose a lightweight CNN model ResNet-56~\cite{he2016deep}, which has only 0.86M parameters. What's more, the inputs of the CNN are low resolution images. With these two designs, we obtain a weak CNN that even performs worse than some VTs. Even if the weak CNN performs poorly, we show that different VTs can perform significantly better than both the weak CNN and the VT baselines in different levels, requiring only a small amount of computational overhead. We provide ablation study to prove that CNNs with different sizes can provide guidance on local information to the VT and help VT make great progress on tiny datasets. In other words, the capability of the CNN will not be the performance bottleneck, reflecting the effectiveness and the efficiency of our framework.

\section{Experiments}

In this section, we demonstrate the effectiveness and efficiency of our approach on image classification task. Firstly, we evaluate different VTs' performance on various datasets with and without our method, and compare the method with two other similar ones. Then, we explore the effect of our method via visualization same as in Section~\ref{section:introduction}. At last, we provide ablation studies to discuss the design of our method.

\subsection{Main Results}

\noindent \textbf{Datasets}
We evaluate our method on CIFAR-100~\cite{nilsback2008automated} dataset (with 50,000 training samples and  10,000 test samples for 100 classes) and Oxford Flowers~\cite{krizhevsky2009learning} dataset (with 2,040 training samples and  6,149 test samples for 102 classes) of natural image domain. Furthermore, we also explore its performance on Chaoyang~\cite{zhu2021hard} dataset (with 4021 training samples and 2139 test samples for 4 classes) of medical image domain, in which large-scale datasets and pre-trained models are hard to obtain, making it a practical application domain for our method.
\smallskip

\noindent \textbf{Models}
To illustrate the generality, we test our method for different kinds of VTs including pure transformer architectures (DeiT~\cite{touvron2021training}, T2T~\cite{yuan2021tokens}), hierarchical architectures (PVT~\cite{wang2021pyramid}, PiT~\cite{heo2021rethinking}), and architectures with convolutional inductive bias (PVTv2~\cite{wang2021pvtv2}, ConViT~\cite{d2021convit}).

\noindent \textbf{Implementation Details}
We adopt the training settings used by Liu et al. \cite{liu2021efficient} for all VTs. Specifically, we employ the AdamW~\cite{loshchilov2017decoupled} optimizer with an initial learning rate of 5e-4 and a weight decay of 0.05. The learning rate is finally reduced to 5e-6 following the cosine learning rate policy~\cite{loshchilov2016sgdr}. All VTs are trained for 300 epochs (with linear warm-up for 20 epochs) on $ 224 \times 224 $ resolution images if not specified. The hyperparameter $ R $ of our method is fixed to 1.0 for convenience, while $ \beta $ is selected for different VTs, which is discussed in detail through ablation study. Considering the efficiency, we choose ResNet-56~\cite{he2016deep} as the guidance model and train it on $ 32 \times 32 $ resolution images. For the CNN baseline ResNet-18~\cite{he2016deep}, we train it with the same setting of VTs for fair comparison, except that we use the SGD optimizer with an initial learning rate of 0.1 and a weight decay of 5e-4. We choose the smallest variant for all VTs, in order to match the size of the CNN baseline. Further implementation details are provided in supplementary material.
\smallskip

\begin{table}[t]
    \centering
    \caption{\textbf{Results of different VTs and datasets.} As shown here, our method can be generalized to different VTs and datasets, and we make VTs to be effective options even on tiny datasets. What's more, it's worth mentioning that all VTs perform significantly better than the CNN guidance model.}
    \label{tab:4.1}
    \begin{tabular*}{1.0\textwidth}{@{\extracolsep{\fill}}lccc}
        \toprule[1.2pt]
        \multirow{2}{*}{\textbf{Model}} & \multicolumn{3}{c}{\textbf{Top-1 Acc.}} \\
        \cline{2-4}
         & \textbf{CIFAR-100} & \textbf{Flowers} & \textbf{Chaoyang} \\
        \toprule[1.2pt]
        
        \multicolumn{4}{l}{\emph{Guidance Model}} \\
        \hline
        
        ResNet-56\cite{he2016deep} (32 res.) & 70.43 & 59.83 & 78.12 \\
        \hline
        
        \multicolumn{4}{l}{\emph{CNN Baseline}} \\
        \hline
    	
    	ResNet-18\cite{he2016deep} (224 res.) & 79.00 & 69.23 & 84.71 \\
    	\hline
        
        \multicolumn{4}{l}{\emph{Pure Transformer}} \\
        \hline
            	
        DeiT-Ti\cite{touvron2021training} & 65.08 & 50.06 & 82.00 \\
        DeiT-Ti + $ L_{guidance} $ & 78.15(+\textbf{13.07}) & 68.50(+\textbf{18.44}) & 84.20(+2.20) \\

        T2T-ViT-7\cite{yuan2021tokens} & 69.37 & 65.20 & 80.74 \\
        T2T-ViT-7 + $ L_{guidance} $ & 78.35(+8.98) & 68.97(+3.77) & 82.89(+2.15) \\
        \hline

        \multicolumn{4}{l}{\emph{Transformer with Hierarchy Structure}} \\
        \hline

        PiT-Ti\cite{heo2021rethinking} & 73.58 & 56.40 & 82.70 \\
        PiT-Ti + $ L_{guidance} $ & 78.48(+4.90) & 68.32(+11.92) & 83.78(+1.08) \\
        
        PVT-Ti\cite{wang2021pyramid} & 69.22 & 62.32 & 73.68 \\
        PVT-Ti + $ L_{guidance} $ & 77.07(+7.85) & 70.61(+8.29) & \textbf{85.65}(+\textbf{11.97}) \\
        \hline
            	 
        \multicolumn{4}{l}{\emph{Transformer with Convolutional Inductive Bias}} \\
        \hline
        
        PVTv2-B0\cite{wang2021pvtv2} & 77.44 & 67.51 & 82.05 \\
        PVTv2-B0 + $ L_{guidance} $ & \textbf{79.30}(+1.86) & \textbf{72.34}(+4.83) & 84.25(+2.20) \\
        
        ConViT-Ti\cite{d2021convit} & 75.32 & 57.51 & 82.47 \\
        ConViT-Ti + $ L_{guidance} $ & 78.95(+3.63) & 67.04(+9.53) & 84.10(+1.63) \\
    	 
    	\bottomrule[1.2pt]
    \end{tabular*}
\end{table}

\noindent \textbf{Results} Table~\ref{tab:4.1} shows the experimental results of different kinds of VTs on various datasets. We find that our method gets different degrees of improvement for different VTs. The pure transformer models perform the worst since the self-attention mechanism lacks distance limitation and our method brings surprising improvements to these models. The VTs with convolutional inductive bias show not bad performance, and our method can make these strong baselines even better, which shows the potential of using VTs on small datasets to be another choice besides CNN. Meanwhile, it is worth noting that our method can also be generalized to medical image domain, in which training from scratch on small datasets is inevitable. To summarize, our approach improves different VTs by substantial margins on small datasets and makes it possible for VTs to surpass CNNs.

To prove the role of proposed locality guidance in accelerating convergence, we also train these VTs with shorter schedule on CIFAR-100 dataset. The experimental results are given in Table~\ref{tab:100epoch}. Even with only $ 1/3 $ training epochs, our method can largely improve the baseline, demonstrating the efficiency. However, it is reasonable that a bigger improvement can be achieved with more training epochs.

\begin{table}[t]
    \centering
    \caption{\textbf{Results of training for 100/300 epochs.} It is possible to achieve excellent results in shorter training schedule with our method, demonstrating its efficiency.}
    \label{tab:100epoch}
    \begin{tabular*}{0.8\textwidth}{@{\extracolsep{\fill}}cccccc}
        \toprule[1.2pt]
        \multirow{2}{*}{Model} & \multicolumn{5}{c}{Top-1 Acc.}\\
        \cline{2-6}
        & Baseline & \multicolumn{2}{c}{100 Epoches} & \multicolumn{2}{c}{300 Epoch} \\
        \toprule[1.2pt]
        
        DeiT-Tiny & 65.08 & 77.29 & +12.21 & 78.15 & +13.07 \\
        T2T-ViT-7 & 69.37 & 77.16 & +7.79 & 78.35 & +8.98 \\
        PiT-Tiny & 73.58 & 77.61 & +4.03 & 78.48 & +4.90 \\
        PVT-Tiny & 69.22 & 76.20 & +6.98 & 77.07 & +7.85 \\

        \bottomrule[1.2pt]
    \end{tabular*}
\end{table}

\begin{table}[t]
    \centering
    \caption{\textbf{Comparison with the method of Liu et al. \cite{liu2021efficient} (100 epochs).} Our method achieves better performance.}
    \label{tab:compare_liu}
    \begin{tabular*}{0.8\textwidth}{@{\extracolsep{\fill}}cccccc}
        \toprule[1.2pt]
        \multirow{2}{*}{Model} & \multirow{2}{*}{Method} & \multicolumn{4}{c}{Top-1 Acc.} \\
        \cline{3-6}
        &  & \multicolumn{2}{c}{CIFAR-100} & \multicolumn{2}{c}{Flowers} \\ 
        \toprule[1.2pt]
        
        \multirow{3}{*}{T2T-ViT-14} & baseline & 65.16 & - & 31.73 & - \\
        & $ L_{drloc} $\cite{liu2021efficient} & 68.03 & +2.87 & 34.35 & +2.62 \\
        & Ours & \textbf{77.84} & \textbf{+12.68} & \textbf{67.71} & \textbf{+35.98} \\
        \hline
        
        \multirow{3}{*}{CvT-13} & baseline & 73.50 & - & 54.29 & - \\
        & $ L_{drloc} $\cite{liu2021efficient} & 74.51 & +1.01 & 56.29 & +2.00 \\
        & Ours & \textbf{76.55} & \textbf{+3.05} & \textbf{65.13} & \textbf{+10.84} \\
        
        \bottomrule[1.2pt]
    \end{tabular*}
\end{table}

Table~\ref{tab:compare_liu} compares our method with the method of Liu et al.~\cite{liu2021efficient}, which designs an additional self-supervised task parallel with the supervised classification task. Their self-supervised task is defined as predicting the distance in 2D space of any two tokens, aiming to constrain the globality of VTs. We argue that there are two shortcomings of this approach. Firstly, a recent research~\cite{raghu2021vision} points out that ViT highly maintains spatial location information, so this self-supervised task may be too easy for VTs. Secondly, only implementing this task at the last layer of VTs makes it hard for shallow layers to catch information, and thus it will lead to a limited boost. As shown in Table~\ref{tab:compare_liu}, with the hierarchical locality guidance of CNN, our method improves VTs more significantly.

Table~\ref{tab:compare_touvron} compares our method with the method of Touvron et al.~\cite{touvron2021training}, which distills the knowledge of CNNs from logits. Although it is originally used in medium-size datasets, we make a comparison between them since they both introduce knowledge distillation and CNNs. The main difference lies in the aim of introducing knowledge distillation, which provides a kind of guidance in our method and learns the classification results of the CNN in DeiT, respectively. Thus in our method, only a lightweight CNN is required. Comparing the experimental results under the same setting, the performance boost in DeiT is still limited, though the distillation method of DeiT seems effective. With our method, the performance of the VT can surpass the CNN guidance model a lot, proving that the weak CNN guidance model won't be the performance bottleneck for the VT. Besides, additional \emph{distill} token in DeiT, which will increase the computational cost during inference, is not necessary in our method.

\begin{table}[t]
    \centering
    \caption{\textbf{Comparison with Touvron et al. \cite{touvron2021training}.} Our method reaches higher performance.}
    \label{tab:compare_touvron}
    \begin{tabular*}{0.8\textwidth}{@{\extracolsep{\fill}}cccc}
        \hline
        Student Model & Teacher Model & Method & Top-1 Acc. \\
        \hline
        \multirow{3}{*}{\makecell*[c]{DeiT-Tiny\\(65.08)}} & \multirow{3}{*}{\makecell*[c]{ResNet-56\\(70.43)}} & DeiT-Soft & 66.92(+1.84) \\
        & & DeiT-Hard & 73.25(+8.17) \\
        &  & Ours & 78.15(+13.07) \\
        \hline
    \end{tabular*}
\end{table}

\begin{table}[t]
    \centering
    \caption{\textbf{Comparison of attention distance.} The VTs with locality guidance learn to pay more attention to locality than those ones trained from scratch.}
    \label{tab:attention_distance}
    \begin{tabular*}{0.8\textwidth}{@{\extracolsep{\fill}}ccccccc}
        \hline
        Model & DeiT & T2T & PiT & PVT & PVTv2 & ConViT \\
        \hline
        w/o $ L_{guidance} $ & 0.0336 & 0.0338 & 0.0421 & 0.2590 & 0.2622 & 0.0250 \\
        w/ $ L_{guidance} $ & 0.0185 & 0.0181 & 0.0293 & 0.2059 & 0.2568 & 0.0171 \\
        \hline
    \end{tabular*}
\end{table}

\subsection{Discussion}

\begin{figure}[ht]
    \centering
    \includegraphics[width=0.28\textwidth]{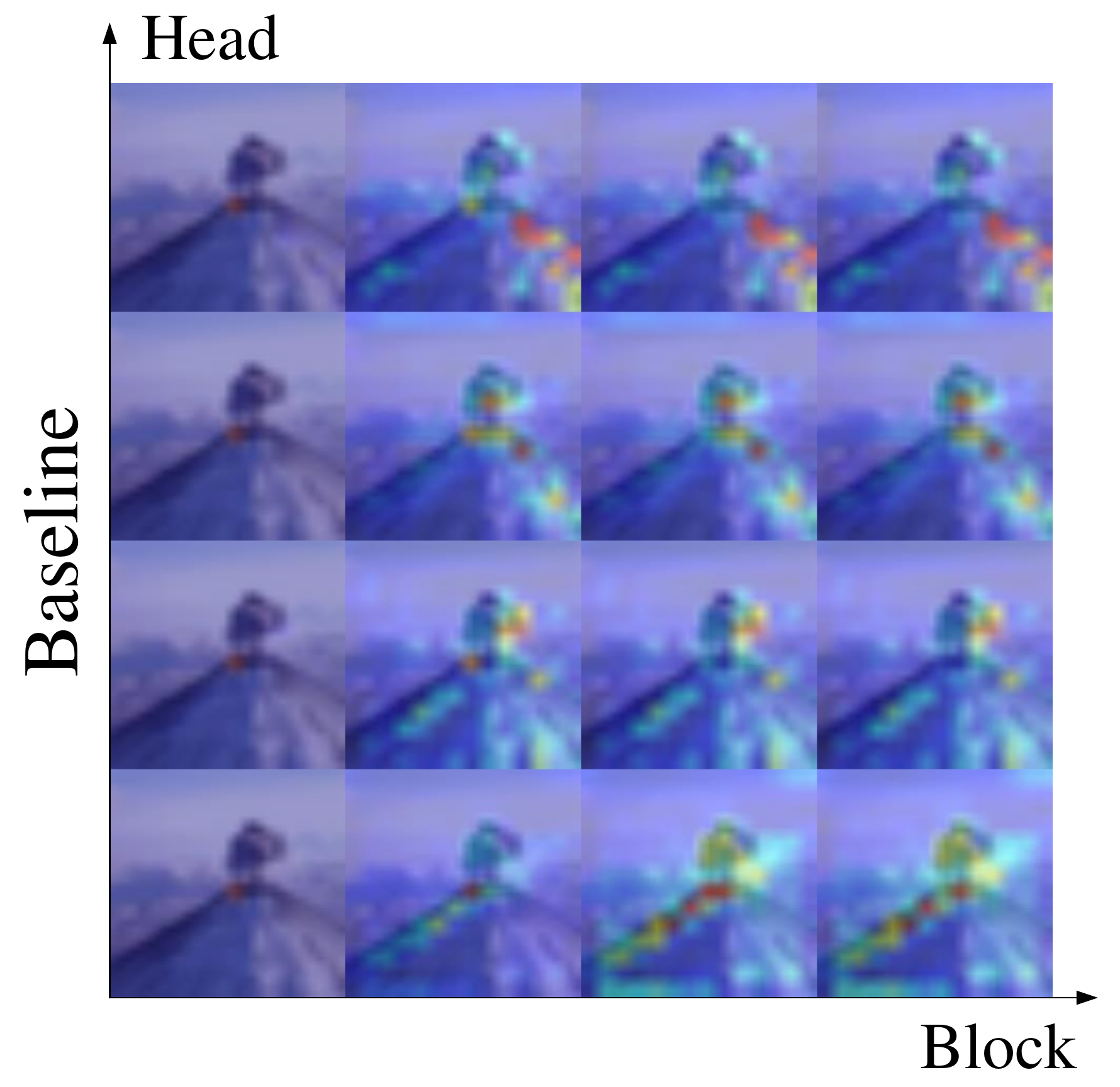}
    \includegraphics[width=0.28\textwidth]{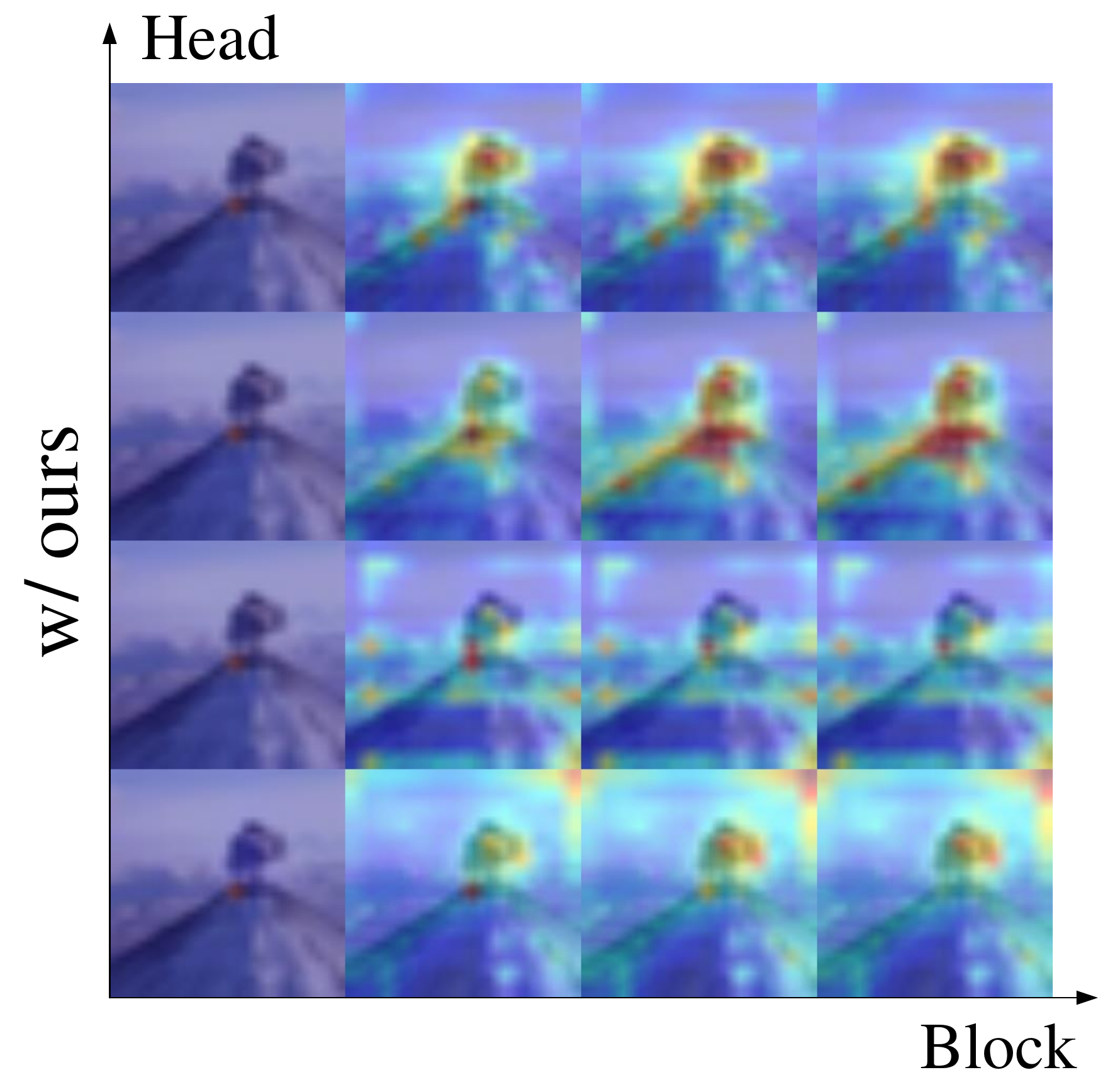}
    \includegraphics[width=0.28\textwidth]{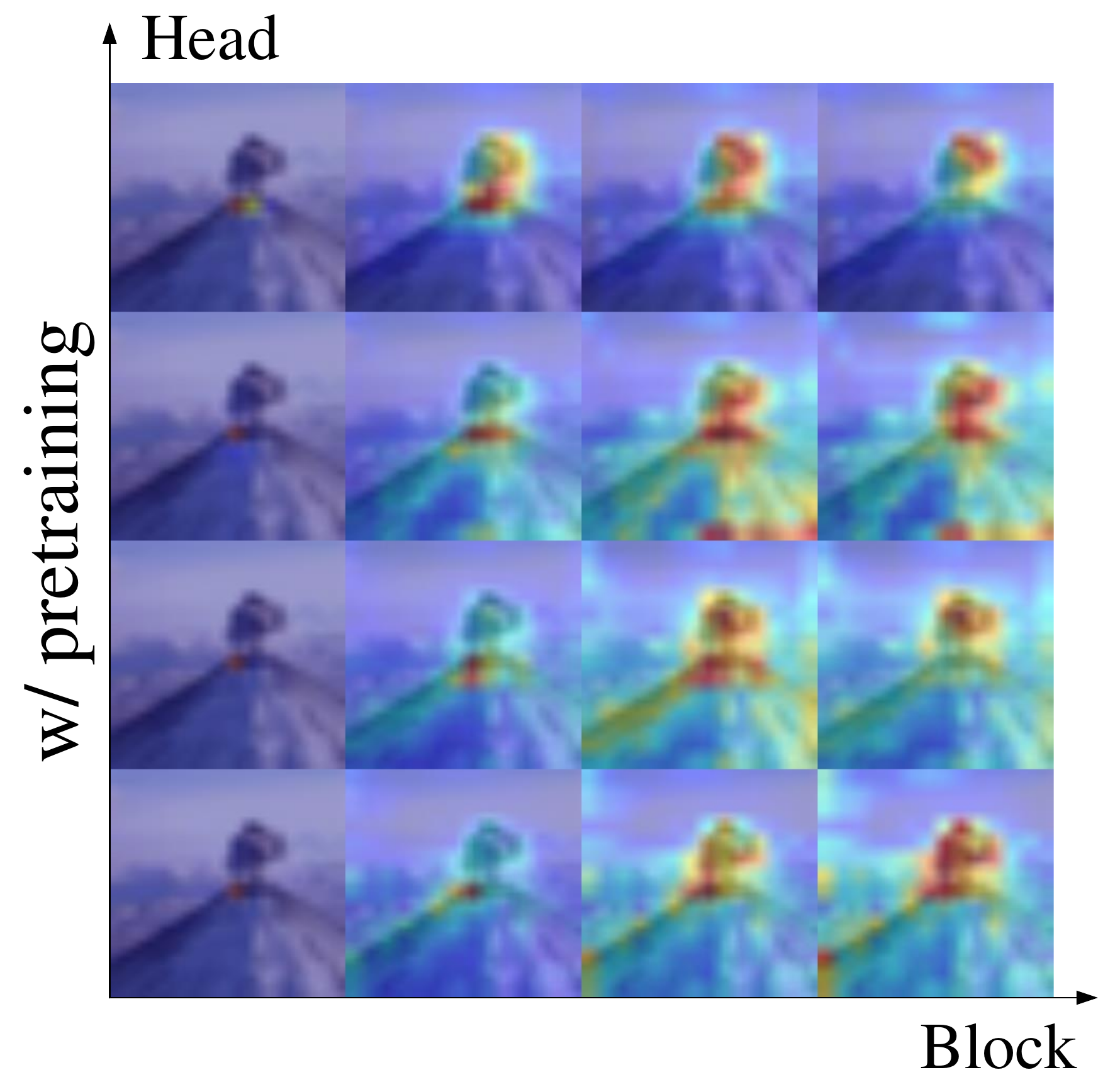}
    \caption{\textbf{Comparison of attention map.} The attention map of VT with locality guidance (center) present similar to the ones of the pre-trained model (right) and are more reasonable than the ones of baseline (left).}
    \label{fig:discussion_map}
\end{figure}

The purpose of our method is to simplify the process of learning locality for VT. To prove that it does achieve such a purpose, we compare the attention statistics with or without our method via the same approaches in Section~\ref{section:introduction}. In addition, we calculate the attention distance averaging on each head and each layer. All the results shown in this section are produced on CIFAR-100 test set. The attention distance given in Table~\ref{tab:attention_distance} shows that the VTs with locality guidance learn to pay more attention to locality than those ones trained from scratch. By checking the attention maps of T2T-ViT-14~\cite{yuan2021tokens} shown in Fig.~\ref{fig:discussion_map}, we find that the VT can learn more meaningful and generalizable information after adding locality guidance and the attended scope is expanded firstly and then focused on region of interests gradually. In summary, our method can play a similar role as pre-training to simplify the learning process of VTs. Moreover, the VT acts in its own way thanks to the dual-task setting. As a result, the proposed method achieves significant improvements for VTs on tiny datasets.

\subsection{Ablation Study}
\label{section:ablation}

To defend the design options in our method, we perform ablation studies on the guidance positions, the hyperparameter $ \beta $ used to balance imitation and self-learning, the channel transformation function and the complexity of the CNN model. All results shown in this section are based on DeiT-Tiny, CIFAR-100 dataset and training schedule of 100 epochs.

Table~\ref{tab:position} shows the influence of different guidance positions. It can be concluded that completely utilizing the features of the CNN is important to achieve remarkable improvement. We also observe that $ R $ is related to the depth of the features from the CNN. For example, it is optimal to set $ R=1.0 $ while all the features from the CNN are selected, and $ R=0.5 $ or $ R=0.75 $ while $ 2/3 $ of the features are selected. This can be interpreted as that the VTs understand images in a hierarchical way similar to CNNs, so that the guidance may be ambiguous when the features from the CNN are misaligned or missing.

To verify the impact of hyperparameter $ \beta $, we adopt different $ \beta $ evenly distributed in $ [0, 3.0] $. The experimental results in Table~\ref{tab:beta} demonstrate the role of $ \beta $ for balancing imitation and self-learning. The imitation signal will be too weak if $ \beta $ is too small, leading to that the VT can not receive enough guidance on locality. Our method can show a significant performance improvement when $ \beta $ is within a suitable range.

We implement different transformation functions to replace the learnable linear projection in Equation~\eqref{equation:linear}, which aligns the channel dimension of features from different models. We test each transformation function with the corresponding optimal $ \beta $. The experimental results given in Table~\ref{tab:transform} show that different transformation functions are feasible under our framework. Although different transformation functions express the information in different ways, they all play a common role to guide the VT to understand image information more easily. Nonetheless, the Attention~\cite{komodakis2017paying} and the Similarity~\cite{tung2019similarity} methods perform not so well, due to the fixed form. The adopted learnable linear projection is more flexible and achieves the largest improvement.

As for the guidance model, we apply three CNNs which have the same architecture but different number of layers. From Table~\ref{tab:teacher} we can find that even though the three CNNs show a huge performance gap, the difference between improvements for VT brought by them is relatively small. This phenomenon reveals that our method acts as a guidance for the VT to learn locality, rather than fully transferring the knowledge of the CNN, which allows our method to become very efficient by using lightweight CNNs.

\begin{table}[t]
    \centering
    \caption{\textbf{Ablation study results on guidance position.} The ratio $ R $ in Equation~\eqref{equation:position} is related to the utilization rate of the CNN. It is optimal to utilize all features from the CNN.}
    \label{tab:position}
    \begin{tabular*}{0.6\textwidth}{@{\extracolsep{\fill}}cccc}
        \hline
        CNN Layers & $ R $ & VT Layers & Top-1 Acc. \\
        \hline
        
        \multirow{4}{*}{(1,2,3)} & 0.25 & (1,2,3) & 70.56 \\
        & 0.50 & (1,3,6) & 75.26 \\
        & 0.75 & (1,5,9) & 76.43 \\
        & 1.00 & (1,6,12) & \textbf{77.29} \\
        \hline
        
        \multirow{4}{*}{(1,2)} & 0.25 & (1,3) & 65.93 \\
        & 0.50 & (1,6) & 67.44 \\
        & 0.75 & (1,9) & \textbf{68.01} \\
        & 1.00 & (1,12) & 67.08 \\
        \hline
    \end{tabular*}
\end{table}

\begin{table}[t]
    \centering
    \begin{minipage}{0.32\textwidth}
        \setlength\tabcolsep{4.5pt}
        \centering
        \caption{Ablation study results on the factor $ \beta $ in Equation~\eqref{equation:total_loss}.}
        \label{tab:beta}
        \begin{tabular}{cc|cc}
            \hline
            $ \beta $ & Acc. & $ \beta $ & Acc.\\
            \hline
            0.0 & 65.08 & 2.0 & 77.00 \\
            0.5 & 70.88 & 2.5 & \textbf{77.29} \\
            1.0 & 74.91 & 3.0 & 77.18 \\
            1.5 & 76.37 &  &  \\
            \hline
        \end{tabular}
    \end{minipage}
    \begin{minipage}{0.25\textwidth}
        \setlength\tabcolsep{7pt}
        \centering
        \caption{Ablation on the transformation in Equation~\eqref{equation:linear}.}
        \label{tab:transform}
        \begin{tabular}{cc}
            \hline
            Method & Acc. \\
            \hline
            None & 65.08 \\
            AT\cite{komodakis2017paying} & 73.51 \\
            SP\cite{tung2019similarity} & 67.36 \\
            Linear & \textbf{77.29} \\
            \hline
        \end{tabular}
    \end{minipage}
    \begin{minipage}{0.41\textwidth}
        \setlength\tabcolsep{2pt}
        \centering
        \caption{Ablation on the complexity of the guidance model by changing the number of layers.}
        \label{tab:teacher}
        \begin{tabular}{ccc}
            \hline
            CNN Model & CNN Acc. & VT Acc. \\
            \hline
            None & - & 65.08 \\
            ResNet-20 & 62.91 & 72.91 \\
            ResNet-56 & 70.43 & \textbf{77.29} \\
            ResNet-110 & \textbf{74.70} & 76.62 \\
            \hline
        \end{tabular}
    \end{minipage}
\end{table}

\section{Conclusion}

In this paper, we introduce an effective and efficient method, which significantly improves the performance of VTs on tiny datasets. It is usually difficult to learn locality in an image for VTs when training from scratch with limited data. To this end, we propose to provide locality guidance by imitating the features learned by a lightweight CNN. Meanwhile, VTs also learn by themselves through supervision to act in a suitable way for them. Extensive experiments confirm the applicability of our method in both natural image domain and medical image domain, as well as for different VTs. We hope that our approach will advance the wider application of transformers on vision tasks, especially for the tiny datasets.

\subsubsection{Acknowledgements}

This work is supported by the Nature Science Foundation of China (No.61972217, No.62081360152, No.62006133), Natural Science Foundation of Guangdong Province in China (No.2019B1515120049, 2020B11113 40056). Li Yuan is supported in part by PKU-Shenzhen Start-Up Research Fund (1270110283).

\clearpage
%
%

\begin{thebibliography}{10}
\providecommand{\url}[1]{\texttt{#1}}
\providecommand{\urlprefix}{URL }
\providecommand{\doi}[1]{https://doi.org/#1}

\bibitem{abnar2020quantifying}
Abnar, S., Zuidema, W.: Quantifying attention flow in transformers. In:
  Proceedings of the 58th Annual Meeting of the Association for Computational
  Linguistics. pp. 4190--4197 (2020)

\bibitem{ahn2019variational}
Ahn, S., Hu, S.X., Damianou, A., Lawrence, N.D., Dai, Z.: Variational
  information distillation for knowledge transfer. In: Proceedings of the
  IEEE/CVF Conference on Computer Vision and Pattern Recognition. pp.
  9163--9171 (2019)

\bibitem{arnab2021vivit}
Arnab, A., Dehghani, M., Heigold, G., Sun, C., Lu{\v{c}}i{\'c}, M., Schmid, C.:
  Vivit: A video vision transformer. In: Proceedings of the IEEE/CVF
  International Conference on Computer Vision. pp. 6836--6846 (2021)

\bibitem{brown2020language}
Brown, T., Mann, B., Ryder, N., Subbiah, M., Kaplan, J.D., Dhariwal, P.,
  Neelakantan, A., Shyam, P., Sastry, G., Askell, A., et~al.: Language models
  are few-shot learners. Advances in neural information processing systems
  \textbf{33},  1877--1901 (2020)

\bibitem{carion2020end}
Carion, N., Massa, F., Synnaeve, G., Usunier, N., Kirillov, A., Zagoruyko, S.:
  End-to-end object detection with transformers. In: European conference on
  computer vision. pp. 213--229. Springer (2020)

\bibitem{changyong2019knowledge}
Changyong, S., Peng, L., Yuan, X., Yanyun, Q., Longquan, D., Lizhuang, M.:
  Knowledge squeezed adversarial network compression. arXiv preprint
  arXiv:1904.05100  (2019)

\bibitem{deng2009imagenet}
Deng, J., Dong, W., Socher, R., Li, L.J., Li, K., Fei-Fei, L.: Imagenet: A
  large-scale hierarchical image database. In: 2009 IEEE conference on computer
  vision and pattern recognition. pp. 248--255. IEEE (2009)

\bibitem{devlin2018bert}
Devlin, J., Chang, M.W., Lee, K., Toutanova, K.: Bert: Pre-training of deep
  bidirectional transformers for language understanding. arXiv preprint
  arXiv:1810.04805  (2018)

\bibitem{dosovitskiy2020image}
Dosovitskiy, A., Beyer, L., Kolesnikov, A., Weissenborn, D., Zhai, X.,
  Unterthiner, T., Dehghani, M., Minderer, M., Heigold, G., Gelly, S., et~al.:
  An image is worth 16x16 words: Transformers for image recognition at scale.
  arXiv preprint arXiv:2010.11929  (2020)

\bibitem{d2021convit}
d’Ascoli, S., Touvron, H., Leavitt, M.L., Morcos, A.S., Biroli, G., Sagun,
  L.: Convit: Improving vision transformers with soft convolutional inductive
  biases. In: International Conference on Machine Learning. pp. 2286--2296.
  PMLR (2021)

\bibitem{gou2021knowledge}
Gou, J., Yu, B., Maybank, S.J., Tao, D.: Knowledge distillation: A survey.
  International Journal of Computer Vision  \textbf{129}(6),  1789--1819 (2021)

\bibitem{hassani2021escaping}
Hassani, A., Walton, S., Shah, N., Abuduweili, A., Li, J., Shi, H.: Escaping
  the big data paradigm with compact transformers. arXiv preprint
  arXiv:2104.05704  (2021)

\bibitem{he2016deep}
He, K., Zhang, X., Ren, S., Sun, J.: Deep residual learning for image
  recognition. In: Proceedings of the IEEE conference on computer vision and
  pattern recognition. pp. 770--778 (2016)

\bibitem{he2019knowledge}
He, T., Shen, C., Tian, Z., Gong, D., Sun, C., Yan, Y.: Knowledge adaptation
  for efficient semantic segmentation. In: Proceedings of the IEEE/CVF
  Conference on Computer Vision and Pattern Recognition. pp. 578--587 (2019)

\bibitem{heo2019knowledge}
Heo, B., Lee, M., Yun, S., Choi, J.Y.: Knowledge transfer via distillation of
  activation boundaries formed by hidden neurons. In: Proceedings of the AAAI
  Conference on Artificial Intelligence. vol.~33, pp. 3779--3787 (2019)

\bibitem{heo2021rethinking}
Heo, B., Yun, S., Han, D., Chun, S., Choe, J., Oh, S.J.: Rethinking spatial
  dimensions of vision transformers. In: Proceedings of the IEEE/CVF
  International Conference on Computer Vision. pp. 11936--11945 (2021)

\bibitem{hinton2015distilling}
Hinton, G., Vinyals, O., Dean, J.: Distilling the knowledge in a neural
  network. arXiv preprint arXiv:1503.02531  (2015)

\bibitem{ke2021chextransfer}
Ke, A., Ellsworth, W., Banerjee, O., Ng, A.Y., Rajpurkar, P.: Chextransfer:
  performance and parameter efficiency of imagenet models for chest x-ray
  interpretation. In: Proceedings of the Conference on Health, Inference, and
  Learning. pp. 116--124 (2021)

\bibitem{kim2017transferring}
Kim, S.W., Kim, H.E.: Transferring knowledge to smaller network with
  class-distance loss  (2017)

\bibitem{komodakis2017paying}
Komodakis, N., Zagoruyko, S.: Paying more attention to attention: improving the
  performance of convolutional neural networks via attention transfer. In: ICLR
  (2017)

\bibitem{krizhevsky2009learning}
Krizhevsky, A., Hinton, G., et~al.: Learning multiple layers of features from
  tiny images  (2009)

\bibitem{lee2011unsupervised}
Lee, H., Grosse, R., Ranganath, R., Ng, A.Y.: Unsupervised learning of
  hierarchical representations with convolutional deep belief networks.
  Communications of the ACM  \textbf{54}(10),  95--103 (2011)

\bibitem{li2022joint}
Li, H., Li, X., Karimi, B., Chen, J., Sun, M.: Joint learning of object graph
  and relation graph for visual question answering. arXiv preprint
  arXiv:2205.04188  (2022)

\bibitem{li2018detnet}
Li, Z., Peng, C., Yu, G., Zhang, X., Deng, Y., Sun, J.: Detnet: Design backbone
  for object detection. In: Proceedings of the European conference on computer
  vision (ECCV). pp. 334--350 (2018)

\bibitem{liu2021efficient}
Liu, Y., Sangineto, E., Bi, W., Sebe, N., Lepri, B., Nadai, M.: Efficient
  training of visual transformers with small datasets. Advances in Neural
  Information Processing Systems  \textbf{34} (2021)

\bibitem{liu2021swin}
Liu, Z., Lin, Y., Cao, Y., Hu, H., Wei, Y., Zhang, Z., Lin, S., Guo, B.: Swin
  transformer: Hierarchical vision transformer using shifted windows. In:
  Proceedings of the IEEE/CVF International Conference on Computer Vision. pp.
  10012--10022 (2021)

\bibitem{loshchilov2016sgdr}
Loshchilov, I., Hutter, F.: Sgdr: Stochastic gradient descent with warm
  restarts. arXiv preprint arXiv:1608.03983  (2016)

\bibitem{loshchilov2017decoupled}
Loshchilov, I., Hutter, F.: Decoupled weight decay regularization. arXiv
  preprint arXiv:1711.05101  (2017)

\bibitem{luo2016face}
Luo, P., Zhu, Z., Liu, Z., Wang, X., Tang, X.: Face model compression by
  distilling knowledge from neurons. In: Thirtieth AAAI conference on
  artificial intelligence (2016)

\bibitem{menze2014multimodal}
Menze, B.H., Jakab, A., Bauer, S., Kalpathy-Cramer, J., Farahani, K., Kirby,
  J., Burren, Y., Porz, N., Slotboom, J., Wiest, R., et~al.: The multimodal
  brain tumor image segmentation benchmark (brats). IEEE transactions on
  medical imaging  \textbf{34}(10),  1993--2024 (2014)

\bibitem{nilsback2008automated}
Nilsback, M.E., Zisserman, A.: Automated flower classification over a large
  number of classes. In: 2008 Sixth Indian Conference on Computer Vision,
  Graphics \& Image Processing. pp. 722--729. IEEE (2008)

\bibitem{passalis2020heterogeneous}
Passalis, N., Tzelepi, M., Tefas, A.: Heterogeneous knowledge distillation
  using information flow modeling. In: Proceedings of the IEEE/CVF Conference
  on Computer Vision and Pattern Recognition. pp. 2339--2348 (2020)

\bibitem{peng2021conformer}
Peng, Z., Huang, W., Gu, S., Xie, L., Wang, Y., Jiao, J., Ye, Q.: Conformer:
  Local features coupling global representations for visual recognition. In:
  Proceedings of the IEEE/CVF International Conference on Computer Vision. pp.
  367--376 (2021)

\bibitem{qiao2021detectors}
Qiao, S., Chen, L.C., Yuille, A.: Detectors: Detecting objects with recursive
  feature pyramid and switchable atrous convolution. In: Proceedings of the
  IEEE/CVF Conference on Computer Vision and Pattern Recognition. pp.
  10213--10224 (2021)

\bibitem{raghu2021vision}
Raghu, M., Unterthiner, T., Kornblith, S., Zhang, C., Dosovitskiy, A.: Do
  vision transformers see like convolutional neural networks? Advances in
  Neural Information Processing Systems  \textbf{34} (2021)

\bibitem{romero2014fitnets}
Romero, A., Ballas, N., Kahou, S.E., Chassang, A., Gatta, C., Bengio, Y.:
  Fitnets: Hints for thin deep nets. arXiv preprint arXiv:1412.6550  (2014)

\bibitem{shen2019meal}
Shen, Z., He, Z., Xue, X.: Meal: Multi-model ensemble via adversarial learning.
  In: Proceedings of the AAAI Conference on Artificial Intelligence. vol.~33,
  pp. 4886--4893 (2019)

\bibitem{touvron2021training}
Touvron, H., Cord, M., Douze, M., Massa, F., Sablayrolles, A., J{\'e}gou, H.:
  Training data-efficient image transformers \& distillation through attention.
  In: International Conference on Machine Learning. pp. 10347--10357. PMLR
  (2021)

\bibitem{tung2019similarity}
Tung, F., Mori, G.: Similarity-preserving knowledge distillation. In:
  Proceedings of the IEEE/CVF International Conference on Computer Vision. pp.
  1365--1374 (2019)

\bibitem{vaswani2017attention}
Vaswani, A., Shazeer, N., Parmar, N., Uszkoreit, J., Jones, L., Gomez, A.N.,
  Kaiser, {\L}., Polosukhin, I.: Attention is all you need. Advances in neural
  information processing systems  \textbf{30} (2017)

\bibitem{wang2019distilling}
Wang, T., Yuan, L., Zhang, X., Feng, J.: Distilling object detectors with
  fine-grained feature imitation. In: Proceedings of the IEEE/CVF Conference on
  Computer Vision and Pattern Recognition. pp. 4933--4942 (2019)

\bibitem{wang2021pyramid}
Wang, W., Xie, E., Li, X., Fan, D.P., Song, K., Liang, D., Lu, T., Luo, P.,
  Shao, L.: Pyramid vision transformer: A versatile backbone for dense
  prediction without convolutions. In: Proceedings of the IEEE/CVF
  International Conference on Computer Vision. pp. 568--578 (2021)

\bibitem{wang2021pvtv2}
Wang, W., Xie, E., Li, X., Fan, D.P., Song, K., Liang, D., Lu, T., Luo, P.,
  Shao, L.: Pvtv2: Improved baselines with pyramid vision transformer.
  Computational Visual Media  \textbf{8}(3),  1--10 (2022)

\bibitem{wu2019distilled}
Wu, A., Zheng, W.S., Guo, X., Lai, J.H.: Distilled person re-identification:
  Towards a more scalable system. In: Proceedings of the IEEE/CVF Conference on
  Computer Vision and Pattern Recognition. pp. 1187--1196 (2019)

\bibitem{wu2021cvt}
Wu, H., Xiao, B., Codella, N., Liu, M., Dai, X., Yuan, L., Zhang, L.: Cvt:
  Introducing convolutions to vision transformers. In: Proceedings of the
  IEEE/CVF International Conference on Computer Vision. pp. 22--31 (2021)

\bibitem{xie2021segformer}
Xie, E., Wang, W., Yu, Z., Anandkumar, A., Alvarez, J.M., Luo, P.: Segformer:
  Simple and efficient design for semantic segmentation with transformers.
  Advances in Neural Information Processing Systems  \textbf{34} (2021)

\bibitem{yang2021tap}
Yang, Z., Lu, Y., Wang, J., Yin, X., Florencio, D., Wang, L., Zhang, C., Zhang,
  L., Luo, J.: Tap: Text-aware pre-training for text-vqa and text-caption. In:
  Proceedings of the IEEE/CVF conference on computer vision and pattern
  recognition. pp. 8751--8761 (2021)

\bibitem{yim2017gift}
Yim, J., Joo, D., Bae, J., Kim, J.: A gift from knowledge distillation: Fast
  optimization, network minimization and transfer learning. In: Proceedings of
  the IEEE Conference on Computer Vision and Pattern Recognition. pp.
  4133--4141 (2017)

\bibitem{yu2019deep}
Yu, Z., Yu, J., Cui, Y., Tao, D., Tian, Q.: Deep modular co-attention networks
  for visual question answering. In: Proceedings of the IEEE/CVF conference on
  computer vision and pattern recognition. pp. 6281--6290 (2019)

\bibitem{yuan2021incorporating}
Yuan, K., Guo, S., Liu, Z., Zhou, A., Yu, F., Wu, W.: Incorporating convolution
  designs into visual transformers. In: Proceedings of the IEEE/CVF
  International Conference on Computer Vision. pp. 579--588 (2021)

\bibitem{yuan2021tokens}
Yuan, L., Chen, Y., Wang, T., Yu, W., Shi, Y., Jiang, Z.H., Tay, F.E., Feng,
  J., Yan, S.: Tokens-to-token vit: Training vision transformers from scratch
  on imagenet. In: Proceedings of the IEEE/CVF International Conference on
  Computer Vision. pp. 558--567 (2021)

\bibitem{yuan2021volo}
Yuan, L., Hou, Q., Jiang, Z., Feng, J., Yan, S.: Volo: Vision outlooker for
  visual recognition. arXiv preprint arXiv:2106.13112  (2021)

\bibitem{zbontar2018fastMRI}
Zbontar, J., Knoll, F., Sriram, A., Murrell, T., Huang, Z., Muckley, M.J.,
  Defazio, A., Stern, R., Johnson, P., Bruno, M., et~al.: fastmri: An open
  dataset and benchmarks for accelerated mri. arXiv preprint arXiv:1811.08839
  (2018)

\bibitem{zhang2021aggregating}
Zhang, Z., Zhang, H., Zhao, L., Chen, T., , Arık, S.O., Pfister, T.: Nested
  hierarchical transformer: Towards accurate, data-efficient and interpretable
  visual understanding. In: AAAI Conference on Artificial Intelligence (AAAI)
  (2022)

\bibitem{zhu2021hard}
Zhu, C., Chen, W., Peng, T., Wang, Y., Jin, M.: Hard sample aware noise robust
  learning for histopathology image classification. IEEE Transactions on
  Medical Imaging  (2021)

\end{thebibliography}

\end{document}